\pdfoutput=1

\documentclass[11pt]{article}

\usepackage[final]{acl}

\usepackage{times}
\usepackage{latexsym}

\usepackage[T1]{fontenc}

\usepackage[utf8]{inputenc}

\usepackage{microtype}

\usepackage{inconsolata}

\usepackage{times}
\usepackage{latexsym}
\usepackage{algorithm}
\usepackage{algorithmic}
\usepackage{graphicx}
\usepackage{booktabs}
\usepackage{algorithm}
\usepackage{algorithmic}
\usepackage{amsfonts,amssymb}
\usepackage{amsmath,bm}
\usepackage{subcaption}
\usepackage{diagbox}
\usepackage{url}
\usepackage{multirow} 
\usepackage{multicol} 
\usepackage{xcolor}
\usepackage{colortbl}
\usepackage{mathrsfs}
\usepackage{tabularx}
\usepackage{makecell}
\usepackage{amsfonts}
\usepackage{bbm}
\usepackage{arydshln}
\usepackage{booktabs}
\usepackage{bbding}
\usepackage{pifont}
\usepackage{tabularx}
\usepackage{float}
\usepackage{soul} 
\usepackage{color, xcolor} 
\title{FuxiMT: Sparsifying Large Language Models for Chinese-Centric Multilingual Machine Translation}

\author{Shaolin Zhu\textsuperscript{1}, Tianyu Dong\textsuperscript{1}, Bo Li\textsuperscript{2}, Deyi Xiong\textsuperscript{1}\footnotemark[1]\\
\textsuperscript{1}College of Intelligence and Computing, Tianjin University, Tianjin, China\\
\textsuperscript{2}School of Software, Tsinghua University, Beijing, China\\
\texttt{\{zhushaolin, tydong, dyxiong\}@tju.edu.cn}\\
\texttt{\{li-b19\}@mails.tsinghua.edu.cn}} 

\begin{document}
\maketitle
\renewcommand{\thefootnote}{\fnsymbol{footnote}}
\footnotetext[1]{Corresponding author.}

\renewcommand{\thefootnote}{\arabic{footnote}}
\begin{abstract}

In this paper, we present FuxiMT, a novel Chinese-centric multilingual machine translation model powered by a sparsified large language model (LLM).
We adopt a two-stage strategy to train FuxiMT. 
We first pre-train the model on a massive Chinese corpus and then conduct multilingual fine-tuning on a large parallel dataset encompassing 65 languages. 
FuxiMT incorporates Mixture-of-Experts (MoEs) and employs a curriculum learning strategy for robust performance across various resource levels. 
Experimental results demonstrate that FuxiMT significantly outperforms strong baselines, including state-of-the-art LLMs and machine translation models, particularly under low-resource scenarios. 
Furthermore, FuxiMT exhibits remarkable zero-shot translation capabilities for unseen language pairs, indicating its potential to bridge communication gaps where parallel data are scarce or unavailable.

\end{abstract}

\section{Introduction}
\label{Introduction}

The rapid advancement of large language models (LLMs) has ushered in a new era for machine translation (MT) \cite{zhu2024multilingual}. 
Despite this, the huge multilingual translation demands for Chinese have not been technologically satisfied, especially in comparison to those for English. 
While LLMs like BLOOM and LLaMA demonstrate multilingual capabilities, their coverage is severely limited. 
LLaMA supports 20 languages, and even the more extensive BLOOM covers only 46. Considering there are over 7,000 languages worldwide, existing LLMs only serve a tiny fraction of the global population, leaving a vast number of people unable to benefit from their multilingual features.


In order to bridge this gap, we propose FuxiMT, a novel Chinese-centric multilingual translation model powered by a sparsified LLM. 
Particularly, FuxiMT, with approximately 13 billion parameters, builds upon the BLOOMz model \cite{workshop2022bloom} and incorporates Sparse Mixture-of-Experts (MoEs) \cite{zhu2025overcoming}. 
This modular structure enables efficient computation and scalability, crucial for handling large models and diverse language data \cite{DBLP:conf/iclr/ShazeerMMDLHD17}.

Our training strategy for FuxiMT leverages the strengths of both monolingual and parallel data through two distinct phases: (1) \textbf{Chinese Pre-training}: We first pre-train the sparse BLOOM model on a massive dataset of 5 billion Chinese tokens. 
This pre-training stage aims to imbue the model with a deep understanding of the Chinese language, ensuring that FuxiMT is inherently Chinese-centric and capable of high-quality Chinese language processing.
(2) \textbf{Multilingual Translation Training}: We then utilize a comprehensive parallel corpus of over 10 billion sentences, covering 65 languages. 
Specifically, we freeze the parameters of the pre-trained BLOOM model and introduce MoE modules at regular intervals within the decoder stack. 
These MoE modules are initialized using a mixed strategy, leveraging both random initialization and weights from corresponding layers of the pre-trained BLOOM model to facilitate knowledge transfer.
We then fine-tune the MoE parameters on the multilingual parallel data, employing a curriculum learning strategy to gradually increase the language coverage and data complexity. 

To summarize, the key contributions of this paper are threefold:
\begin{itemize}
    \item We introduce FuxiMT, a LLM combining BLOOMz with MoE. 
    By dynamically routing inputs to specialized experts, FuxiMT efficiently handles multilingual translation while reducing computational costs, addressing the limitations of English-dominated models.

    \item We develop a Chinese-first pre-training strategy, initializing FuxiMT on 5B Chinese sentences to embed deep Chinese understanding, followed by multilingual fine-tuning on 100B+ parallel sentences across 65 languages. 

    \item Experimental results demonstrate FuxiMT outperforms strong baselines (e.g., NLLB, GPT-3.5). 
    These results highlight its robustness in real-world multilingual translation challenges.
\end{itemize}

\section{Datasets Construction}

The training process for FuxiMT utilizes monolingual and parallel data. 
In this section, we present the construction pipeline from Chinese datasets and the multilingual datasets used by FuxiMT.

\subsection{Chinese Datasets}
\paragraph{Collection.}
We use 3 sources for a high quality post-cleaning pre-training corpus, totaling more than 5 billion sentences with a data size of more than 1 TB.

\textit{Intern WanJuan (260G)}\footnote{https://opendatalab.org.cn/OpenDataLab/WanJuan1\_dot\_0} \cite{he2023wanjuan} is a Chinese corpus constructed by the Shanghai AI Lab. 
It consists of a cleaned pre-training corpus from various sources such as websites, encyclopedias, books, patents, textbooks, exam questions, etc. and is subjected to fine-grained cleaning, deduplication and value alignment.

\textit{WuDaoCorpora (200G)}\footnote{https://data.baai.ac.cn/details/WuDaoCorporaText} is a large, high-quality Chinese dataset constructed by the BAAI. 
It uses more than 20 rules to clean the final dataset from 100 TB of original website data and remove the private data information, and it contains data from more than 50 domains such as education, science and technology.

\textit{ChineseWebText (530G)}\footnote{https://huggingface.co/datasets/CASIA-LM/ChineseWebText}  \cite{chen2023chinesewebtext} is a monolingual Chinese quality dataset extracted from CommonCrawl data and filtered based on manually formulated rules (including data length, sensitive words, proportion of Chinese characters, etc.). 
The quality of the filtered data is evaluated using a BERT-based evaluation model.
\paragraph{Filtering.}
We prepare the above raw datasets to ensure their quality for the training of machine translation models. First, sentences are extracted from paragraphs using predefined rules. 
Sentences shorter than 50 characters or longer than 250 characters are filtered out, leaving sentences of appropriate length for effective training and semantic completeness.
Next, language identification is performed to exclude sentences without Chinese characters, resulting in a dataset consisting solely of Chinese text. 
This step mitigates the effects of noise on the model training.
Finally, characters are normalized, leaving only Chinese characters, English letters, common punctuation marks and spaces. 
This process removes superfluous special characters and formatting symbols, ensuring the purity of the data and improving the effectiveness of model training.

\begin{figure}[t]
    \centering
    \includegraphics[scale=0.43]{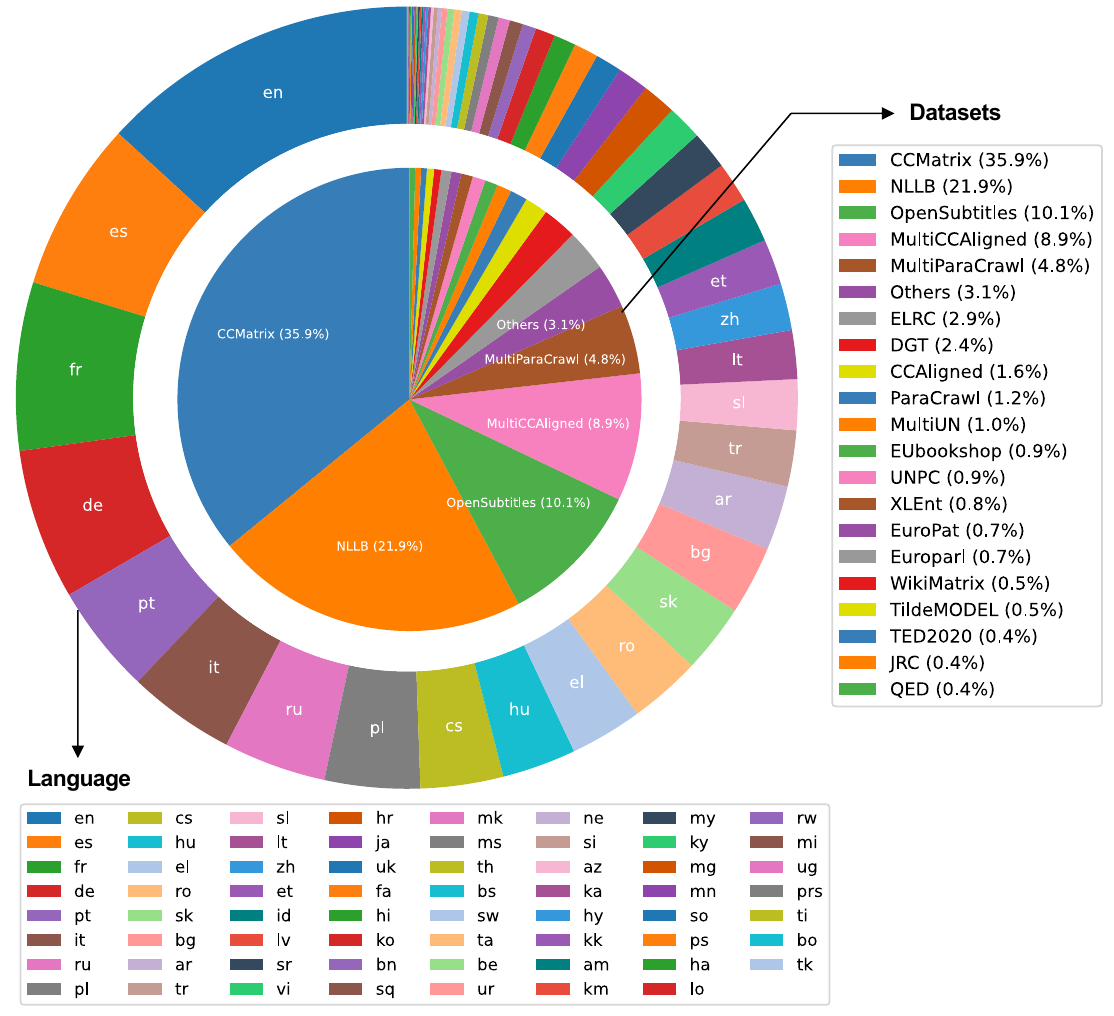}
    \caption{Language and data distribution in the pre-training data of FuxiMT.}
    \label{fig:sft}
\end{figure}


\subsection{Multilingual Datasets}

\paragraph{Collection.}
We use OpusTools\footnote{https://github.com/Helsinki-NLP/OpusTools} \cite{aulamo-etal-2020-opustools} to extract resources from the OPUS project \cite{nygaard2003opus}, a renowned platform for parallel corpora, and create a multilingual dataset.
Specifically, we collect the parallel corpora from prominent projects within OPUS, including NLLB \cite{fan2021beyond}, CCMatrix \cite{schwenk2019ccmatrix} and OpenSubtitles \cite{lison2016opensubtitles2016}.
This comprehensive data collection process results in a corpus of more than 2T, covering 65 languages and over 1900 language pairs (As in Appendix Table \ref{tab:language-codes}).
The dataset comprises over 80 datasets spread across 24,000 subset.
Within this extensive collection, there is 650 subset dedicated to Chinese, totaling 60G.

\paragraph{Preprocessing.}
High-quality parallel corpora are essential for training powerful NMT models. 
However, raw data, such as data crawled from the web, often contains a significant amount of noise, including length inconsistencies, irrelevant information, and sensitive content, which can negatively impact model performance. 
To address this issue, we refer to the preprocessing pipeline proposed by \citet{fan2021beyond}, and use a six-stage cleaning pipeline to create high-quality multilingual parallel corpora.

\textit{Text extraction and preprocessing.} 
We obtain compressed files with the target language pair from the OPUS corpus. 
After decompression, we remove all files that are not in Moses format and keep only the customized plain text files for the target language pair. 
For example, for English-Chinese translation, we only keep files with the suffixes `.en' and `.zh'.

\textit{Proportion of characters.} 
In this step, sentences containing excessive noise or non-target language content are identified and removed. 
Three strategies are used: 
(1) Punctuation ratio filtering: sentences with a punctuation ratio of more than 50\% are removed as they often contain no meaningful information. 
(2) Rule-based filtering: We use the simple\_cleaning script from the preprocess\footnote{https://github.com/kpu/preprocess} toolkit to remove sentences that contain only spaces, excessive non-printable characters, invalid UTF8-encoded characters or excessively long tokens (e.g. DNA sequences, nonsensical repetitions).
(3) Filtering the character ratio of the target language: The clean\_histogram script from the fairseq\footnote{https://github.com/facebookresearch/fairseq/} toolkit identifies and removes sentences with a high proportion of characters outside the target language.
This ensures that the sentences consist mainly of the target language and that irrelevant information is minimized.

\textit{Data length.}
Excessively long or short sentence pairs can interfere with the training of the NMT model. Therefore, we employ several measures to control sentence length: 
(1) We use the SentencePiece Model (SPM) \cite{kudo-richardson-2018-sentencepiece} from the fairseq toolkit to tokenize both the source and the target language. 
(2) Sentence length ratio filtering: The clean-corpus-n.perl script from the moses-smt\footnote{https://github.com/moses-smt/} is used to eliminate sentence pairs with significant length differences, e.g. those where the target sentence is more than three times as long as the source sentence. 
(3) Filtering short text: Documents with an average line length of less than 10 words or a total length of more than 250 characters are discarded as they do not provide enough context for effective model learning.

\textit{Sensitive words.}
To prevent the model from generating harmful or offensive language, we define a list of sensitive words and analyze their frequency in the text. 
If a sentence contains sensitive words whose frequency exceeds a predefined threshold (e.g. 0.5), it is removed from the training dataset.

\textit{Duplication.} 
Duplicate data can lead to overfitting of the model and reduce the generalization ability. 
We use the dedup script from the preprocessing toolkit to identify and remove duplicate record pairs in the parallel data, retaining only the first instance.

\textit{Normalization.}
Finally, we normalize the text formatting to make sure that the same symbols, such as punctuation marks, numbers, and spaces, with the same UTF8 codes, are used in all the datasets.
We use the normalize-punctuation.perl script of the sacremoses toolkit\footnote{https://github.com/alvations/sacremoses}. In addition, to reduce the vocabulary and increase the efficiency of the model, we normalize these different quote marks to the same one.

\begin{figure*}[t]
    \centering
    \includegraphics[scale=0.3]{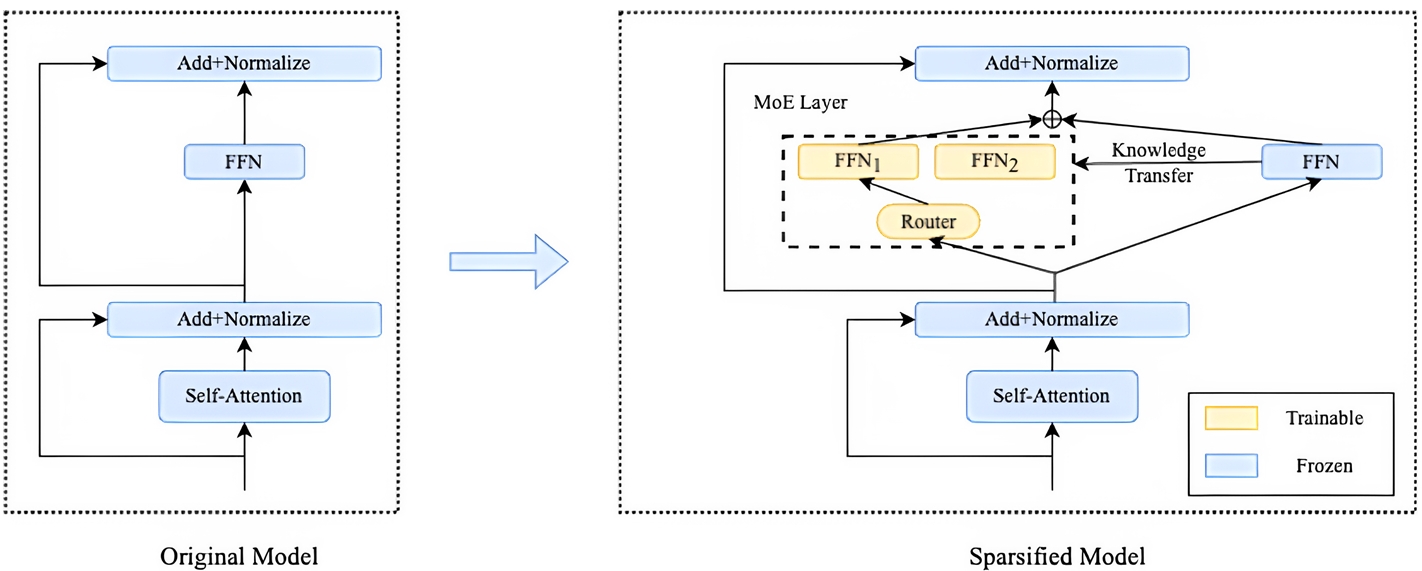}
    \caption{Diagram of FuxiMT. 
    FuxiMT is built upon BLOOMz-7B and fine-tuned on translation and general tasks.}
    \label{fig:model}
\end{figure*}

Significantly, all filtering steps should be applied to both source and target languages, e.g., if the source segment is empty and should be deleted, the target segment should also be deleted to keep the data parallel.
The distribution of language and dataset after preprocessing is shown in Figure \ref{fig:sft}.

\section{FuxiMT}

\subsection{Model Details}

FuxiMT builds upon the BLOOMz model (specifically BLOOMz-7B according to the diagram), a decoder-only LLM. 
As illustrated in Figure \ref{fig:model}, the left part is the ``Original Model'' which is the pre-trained BLOOMz model. 
Crucially, this base model's components (self-attention, Add \& Normalize, and FFN layer) are all marked as ``Frozen''.
This design choice is fundamental to FuxiMT, preventing catastrophic forgetting of the pre-trained linguistic knowledge during subsequent fine-tuning.
The frozen BLOOM model serves as a robust backbone, providing general language understanding capabilities.
The right part of Figure \ref{fig:model} depicts the integrated ``Sparsified Mode'' with the MoE layer. 
The diagram shows the MoE layer being inserted between the ``Add \& Normalize'' and the FFN layer of the original BLOOM model. 
Within the MoE layer, we see multiple FFN blocks labeled ``FFN1'', ``FFN2'' etc. 
These represent the expert networks within the MoE, each specializing in a different aspect of the translation task. 
The ``Router'' component is also depicted within the MoE layer. 
This router is the gating network, responsible for dynamically selecting the most relevant expert for each input token based on its context. 
The diagram shows the router receiving input and then directing the flow of information to the selected expert(s).

\subsection{Post-training}
To effectively integrate the MoE architecture into the BLOOM for FuxiMT, we adjust the MoE-related parameters in the initial configuration as \citet{DBLP:journals/jmlr/FedusZS22}. 
Specifically, we set $sparse\_step$ to 8, which denotes that only one expert is active for every 8 layers in the model.
This sparsity strategy significantly improves computational efficiency while preserving the expressiveness of the model.
$moe\_expert\_count$ is set to 8, which means that eight experts are used in each MoE layer, allowing the model to specialize and distribute the processing of different linguistic data.
For details on other training hyperparameters, see Section \ref{lab:settings}.

The MoE model is formally defined as:
\begin{equation}
P(y_t | \mathbf{X}, \bm{\theta}) = \sum_{i=1}^{k} g_i(\mathbf{X}; \bm{\theta}_g) \cdot P(y_t | \mathbf{X}, \bm{\theta}_{e_i}),
\end{equation}
where $g_i(\mathbf{X}; \bm{\theta}_g)$ represents the gating network responsible for selecting the $i$-th expert $e_i$, and $P(y_t | \mathbf{X}, \bm{\theta}_{e_i})$ denotes the probability distribution over the next token $y_t$ given the initial set $\mathbf{X}$ and the parameters of the selected expert's parameters $ \bm{\theta}_{e_i} $.

\textbf{Chinese pre-training stage:}

First, we train the model using a Causal Language Modeling (CLM) on a 5B monolingual Chinese sentences to improve its understanding of the syntactic and semantic structures of the Chinese language. 
This task is about predicting the next token in a sequence based on the previous tokens. 
Formally, the model learns to maximize the probability of a sequence of tokens $\mathbf{x} = {x_1, x_2, \dots, x_T}$:
\begin{equation}
P(\mathbf{x}) = \prod_{t=1}^{T} P(x_t | \mathbf{x}_{<t}),
\end{equation}
where $ P(x_t | \mathbf{x}_{<t}) $ represents the conditional probability of token $ x_t $ given all preceding tokens $ \mathbf{x}_{<t} = {x_1, x_2, \dots, x_{t-1}} $.
Compared to Masked Language Modeling (MLM), CLM is more suitable for generation-based tasks, which aligns with the primary focus of this stage. 
This advantage is particularly relevant for Chinese, where tokenization plays a significant role in capturing sequential dependencies. 
During training, we use the cross-entropy loss as the optimization objective:
\begin{equation}
\mathcal{L}_{\text{CLM}}(\theta) =-\sum_{t=1}^{T} \log P_\theta(x_t | \mathbf{x}_{<t}),
\end{equation}
where $ P_\theta(x_t | \mathbf{x}_{<t}) $ is the model's predicted probability of token $x_t$ given the preceding tokens, parameterized by $\theta $. The training process aims to minimize this loss across the entire training corpus.

\textbf{Multilingual translation training stage:}

In this stage, we fine-tune the BLOOM MoE model using an extensive multilingual parallel corpus for translation. 
The training data includes over 100 billion sentence pairs in 65 languages, mostly from the OPUS project, covering both high and low resource languages.
We use instruction-tuning strategy to reformulate each translation example into an instruction-driven query.
We create 40 structured instruction templates that guide the model to produce more accurate translations by clearly indicating the context of the source language, the target language, and the nature of the task.
Detailed instruction templates refer to Appendix Table \ref{tab:mt-instructions}.

We use curriculum learning to avoid catastrophic forgetting and ensure robust translation in all languages. 
Training starts with high-resource language pairs and gradually incorporates low-resource languages to prevent over-matching and promote balanced performance. 
We achieve this by weighting the contribution of each language pair to the overall training loss:

\begin{equation}
\mathcal{L}_{\text{total}} = \sum_{t=1}^{T} w_t \cdot \mathcal{L}(X^\text{src}_t, Y^\text{tgt}_t),
\end{equation}
where $w_t$ represents the weighting assigned to the language pair $t$, whereby the importance of the low-resource languages is gradually increased as training progresses.

To further increase the robustness and coverage of our model, especially for low-resource languages, we employ a data augmentation technique known as back-translation.
In this method, the target sentence $Y^\text{tgt}$ is translated back into the source language, resulting in a synthetic source sentence. 
This technique is particularly effective for augmenting data in low-resource language pairs.
The augmented dataset $ D' $ is generated as follows:
\begin{equation}
\begin{split}
D' = & D \cup \{(Y^\text{tgt}, \text{back-translated}(Y^\text{tgt})) | \\
& (X^\text{src}, Y^\text{tgt}) \in D\},
\end{split}
\end{equation}
where $D$ represents the original dataset of sentence pairs and $\text{back-translated}(Y^\text{tgt}) $ is the synthetic source sentence generated by translating the target sentence $Y^\text{tgt} $ into the source language.

\section{Experiments}

\subsection{Settings}
\label{lab:settings}


\begin{table*}[t]
\centering
\begin{tabular}{ccccc}
\toprule
Model & High Resource & Medium Resource & Low Resource & Very Low Resource \\
\midrule
FuxiMT & 37.0257 & 25.2682 & 20.6446 & 20.6649 \\
GPT-3.5 & 27.2460 & 20.0110 & 10.4000 & 9.1491 \\
BLOOMz & 13.6740 & 12.4014 & 14.0044 & 12.5400 \\
NLLB & 18.3420 & 19.4779 & 19.3294 & 20.1124 \\
LLaMAX3-8B & 28.0430 & 26.4063 & 21.9199 & 15.1937 \\
LLaMA-3.1-8B & 2.5763 & 1.9268 & 3.0245 & 2.9714 \\
Mistral-7B-v0.3 & 3.5789 & 6.0708 & 2.4043 & 1.2762 \\
Qwen2.5-7B & 31.6810 & 26.7201 & 13.4558 & 9.8152 \\
Aya-23-8B & 28.9705 & 20.7915 & 6.8899 & 6.3070 \\
Gemma-2-9B & 20.4730 & 16.0337 & 8.8800 & 6.5642 \\
InternLM2.5-7B-chat & 29.6108 & 21.1705 & 7.2183 & 7.7571 \\
\bottomrule
\end{tabular}
\caption{BLEU scores for different resource languages xx-zh.}
\label{tab:model-performance}
\end{table*}

We conducted our experiments on 8 NVIDIA TESLA A100-80G GPUs with Pytorch. 
To mitigate memory consumption and further improve training efficiency, we leverage ZeRO-2 \cite{zero} and Flash-Attention V2 \cite{flashattention-2} technologies. 
For optimization, the standard AdamW optimizer \cite{adamw} was utilized with hyper-parameters set to $\beta_1=0.9$, $\beta_2=0.99$, and $\epsilon=10^{-8}$. 
Additionally, we employ the cosine learning rate scheduler, adopt a warm-up strategy for the learning rate, gradually increasing it to the peak value over the first 1000 steps and then linearly decaying it to zero throughout the remaining training steps.
Detailed training parameters and configurations are provided in Appendix Table \ref{training-details}.

We compare FuxiMT against several strong baselines: BLOOMz-7B \cite{workshop2022bloom}, GPT3.5 \cite{brown2020language}, NLLB \cite{costa2022no}, LLaMAX3-8B \cite{lu2024llamax}, LLaMA-3.1-8B\cite{grattafiori2024llama}, Mistral-7B-v0.3\cite{jiang2023mistral7b}, Qwen 2.5-7B \cite{qwen2.5}, Aya-23-8B \cite{aryabumi2024aya}, Gemma-2-9B \cite{gemma_2024} and InternLM 2.5-7B-chat \cite{cai2024internlm2}.

\subsection{Main Results}

Table \ref{tab:model-performance} presents the performance of FuxiMT and the baselines across different resource levels of languages, categorized as High, Medium, Low, and Very Low resource in terms of the amount of parallel data available (as described in Appendix Table \ref{tab:language-resource}). 

A key observation is the consistent improvement of FuxiMT over the original BLOOMz-7B model. 
This improvement underscores the effectiveness of our proposed methodology, combining MoEs with the two-stage training strategy focusing first on Chinese pre-training and then multilingual fine-tuning.  
The significant gains under the low-resource setting highlight FuxiMT's ability to leverage cross-lingual knowledge transfer, effectively addressing the data scarcity challenges prominent in many languages.
Compared to other LLMs, FuxiMT exhibits competitive performance.  
While some models, like InternLM2.5 and Qwen2.5-7B, show higher scores in high-resource scenarios, FuxiMT maintains a strong advantage under the low and very low resource settings. 
The specialized Chinese pre-training of FuxiMT likely contributes to its superior performance with Chinese-involved translation pairs, even surpassing strong baselines like GPT-3.5 in certain cases.

While NLLB, specifically trained for low-resource translation, performs well in those categories, FuxiMT often achieves comparable or even better results, demonstrating its effectiveness in leveraging monolingual data and the benefits of the MoE architecture.  
The strong performance of LLaMa models, particularly LLaMA 3-8B, highlights the potential of decoder-only models for multilingual translation. 
However, FuxiMT’s tailored training strategy and incorporation of MoEs appear to provide a further edge, especially in data-sparse scenarios.

\subsection{Ablation Study}
To validate the necessity of the sparse MoE architecture, different FFN initialization methods, and the impact of curriculum learning in FuxiMT, we conducted comprehensive ablation experiments by comparing four variants: FuxiMT-Random-Init, FuxiMT-Reuse-Init, FuxiMT-Random-Train, FuxiMT-Order-Train. 
These settings are detailed in Appendix \ref{Detail Setting of Ablation Study}.

\begin{table}[t]
\centering
\begin{tabular}{cc}
\hline
Method & Average BLEU \\
\hline
BLOOMz & 13.08 \\
FuxiMT-Random-Init & 24.46 \\
FuxiMT-Reuse-Init & 25.37 \\
FuxiMT-Random-Train & 19.27 \\
FuxiMT-Order-Train & 21.93 \\
FuxiMT & 26.15 \\
\hline
\end{tabular}
\caption{Ablation Results (Average BLEU across various language pairs)}
\label{tab:ablation_results}
\end{table}



Results are presented in Table \ref{tab:ablation_results}. 
We can find that the FuxiMT configuration, incorporating all designed elements, achieves superior performance. 
The performance difference between FuxiMT and FuxiMT-Random-Init suggests that the pre-training on the Chinese dataset provide much knowledge for the later training. Specifically, the sparse MoE architecture proves crucial for dynamically routing inputs to specialized experts, enabling effective partitioning of the complex multilingual knowledge space. 
We also show that initialization methods has a large influence on the performance of the model.

We then examine the training strategy variants further illuminating the benefits of our curriculum learning approach. 
FuxiMT-Random-Train, which dispenses with curriculum learning in favor of uniform language pair mixing, suffers significant performance degradation. 
This directly underscores the importance of the gradual introduction of languages, particularly the low-resource ones, in mitigating catastrophic forgetting. 
We also find that training the models with FuxiMT-Order-Train can also bring some improvement, which is better than the models with FuxiMT-Random-Train, but still worst than the complete model. 
These findings validate the designed process of MoE initialization, the Chinese-centric pre-training, and curriculum learning.

\section{Conclusion}

In this paper, we present FuxiMT, a multilingual Chinese-centric translation LLM. 
We design a two-stage training approach that explicitly combines Chinese pre-training followed by multilingual fine-tuning with curriculum learning and strategically incorporates MoEs, enabling FuxiMT to excel particularly in low-resource scenarios.
Our results demonstrate substantially superior performance compared to strong baselines. 
Moreover, FuxiMT's remarkable zero-shot translation capabilities further cement its potential as a valuable MT tool for bridging communication gaps across diverse linguistic landscapes where parallel data are scarce.

\bibliography{main}


\appendix
\section{Appendix}
\begin{table}[H]
\flushright
\centering
\begin{tabular}{ll}
\toprule
Hidden Size         & 4,096  \\
Intermediate Size   & 16,384 \\
Heads               & 32    \\
Layers              & 30    \\
Position Embed & 4,096   \\
Vocab Size      & 250,752 \\
MoE Expert Count &  8 \\
Sparse Step & 8 \\
Learning Rate   & 5e-5\\
Batch Size      & 8 \\
Context Length  & 4,096  \\
Weight Decay & 3e-7 \\
FlashAttn V2    &  $\checkmark$ \\
ZeRO-2 & $\checkmark$ \\
\bottomrule
\end{tabular}
\caption{Hyper-parameters of training.}
\label{training-details}
\end{table}

\subsection{Language Pairs Classifications}
Languages can be categorized into high-resource, medium-resource, low-resource, and very low-resource levels, focusing on $xx\rightarrow zh$ translation pairs. FuxiMT supports a total of 65 languages, including Chinese, which results in 64 language pairs. High-resource languages (e.g., English, Spanish, French) have abundant data, medium-resource languages (e.g., Turkish, Lithuanian, Estonian) have moderate data, low-resource languages (e.g., Amharic, Zulu, Nepali) have limited data, and very low-resource languages (e.g., many indigenous languages) have scarce data. The FuxiMT model has achieved competitive translation quality across all these language pairs, demonstrating its effectiveness at different resource levels.

\subsection{Post-training data}

In the multilingual translation training phase, FuxiMT prioritizes Chinese-centric data by allocating over 50\% of its parallel corpus to xx→zh language pairs. This deliberate imbalance ensures the model retains strong Chinese capabilities while acquiring multilingual expertise through curriculum learning (progressively integrating low-resource languages), parameter-efficient MoE architecture (isolating Chinese knowledge in the frozen BLOOM backbone), and back-translation augmentation (increasing Chinese data via synthetic source texts).
\subsection{Baselines}

We compare FuxiMT against several strong baselines reflecting the state-of-the-art in multilingual machine translation.
\begin{itemize}
    \item BLOOMz-7B: The original BLOOMz-7B model serves as a crucial baseline, allowing us to assess the impact of incorporating MoEs and our specialized training regime. 
    This comparison helps isolate the contributions of our proposed modifications.
    \item GPT-3.5: A widely-used large language model known for its strong translation capabilities.
    \item NLLB: A prominent multilingual machine translation model developed by Meta AI, specifically designed for low-resource languages. This comparison highlights FuxiMT's performance relative to a model explicitly trained for translation between diverse languages.We use the 54.5B parameter version of NLLB-200.
    \item LLaMA-3.1-8B: Representative large language models developed by Meta AI, offering a point of comparison against other decoder-only models with different scales and architectures.
    \item LLaMAX3-8B/Mistral-7B-v0.3/Qwen2.5-7B/Aya-23-8B/Gemma-2-9B/InternLM2.5-7B-chat: Other state-of-the-art large language models, providing a diverse range of architectural choices and training methodologies to benchmark against.  
\end{itemize}

\subsection{Results}
In Tables \ref{bleu} and \ref{chrf}, we provide a detailed comparison of FuxiMT's performance with other large language models (LLMs) on 64 language pairs, showcasing both BLEU and CHRF scores. As observed, FuxiMT consistently outperforms other models in translating a majority of these language pairs, with a particularly remarkable advantage in translating from low-resource and very low-resource languages into Chinese. This demonstrates FuxiMT's strong capacity for cross-lingual knowledge transfer, especially in scenarios where parallel data is scarce.

\subsection{Detail Setting of Ablation Study}
\begin{itemize}
\item FuxiMT-Random-Init: All FFN experts are randomly initialized, with curriculum learning applied.
\item FuxiMT-Reuse-Init: All FFN experts reuse the parameters of the corresponding FFN layers in the pre-trained BLOOMz model, along with curriculum learning.
\item FuxiMT-Random-Train: Uses the mixed FFN initialization method but with random training, where all languages are mixed uniformly without weighting.
\item FuxiMT-Order-Train: Applies the mixed FFN initialization method and fixed-order training. Languages are ordered by resource level (high $\rightarrow$ low) but with equal weights.
\end{itemize}
\label{Detail Setting of Ablation Study}
\subsection{Zero-Shot Translation}

\begin{figure*}[t]
    \centering
    \begin{subfigure}[b]{0.44\textwidth}
        \includegraphics[width=\textwidth]{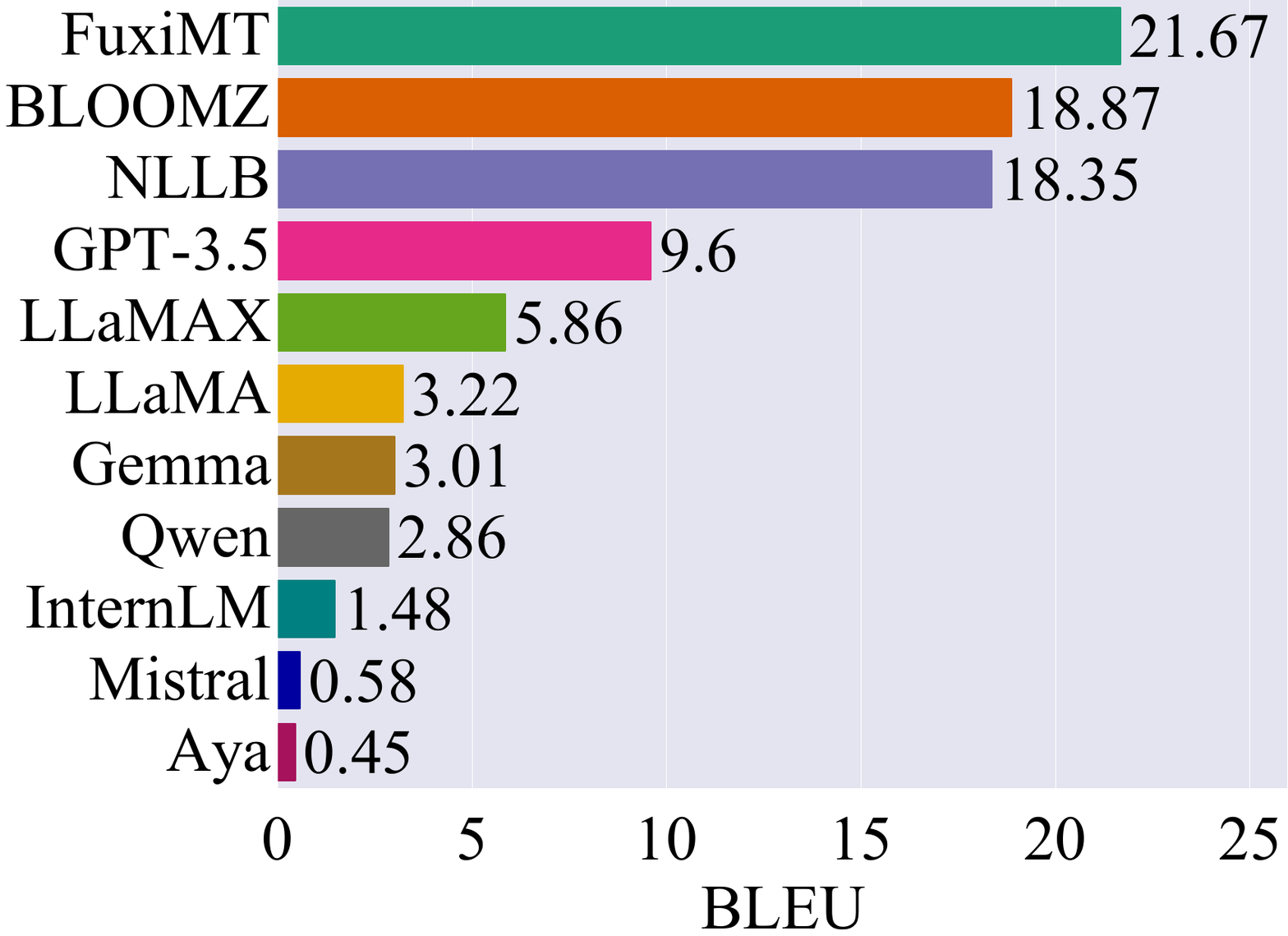}
        \caption{Tigranian}
    \end{subfigure}
    \begin{subfigure}[b]{0.44\textwidth}
        \includegraphics[width=\textwidth]{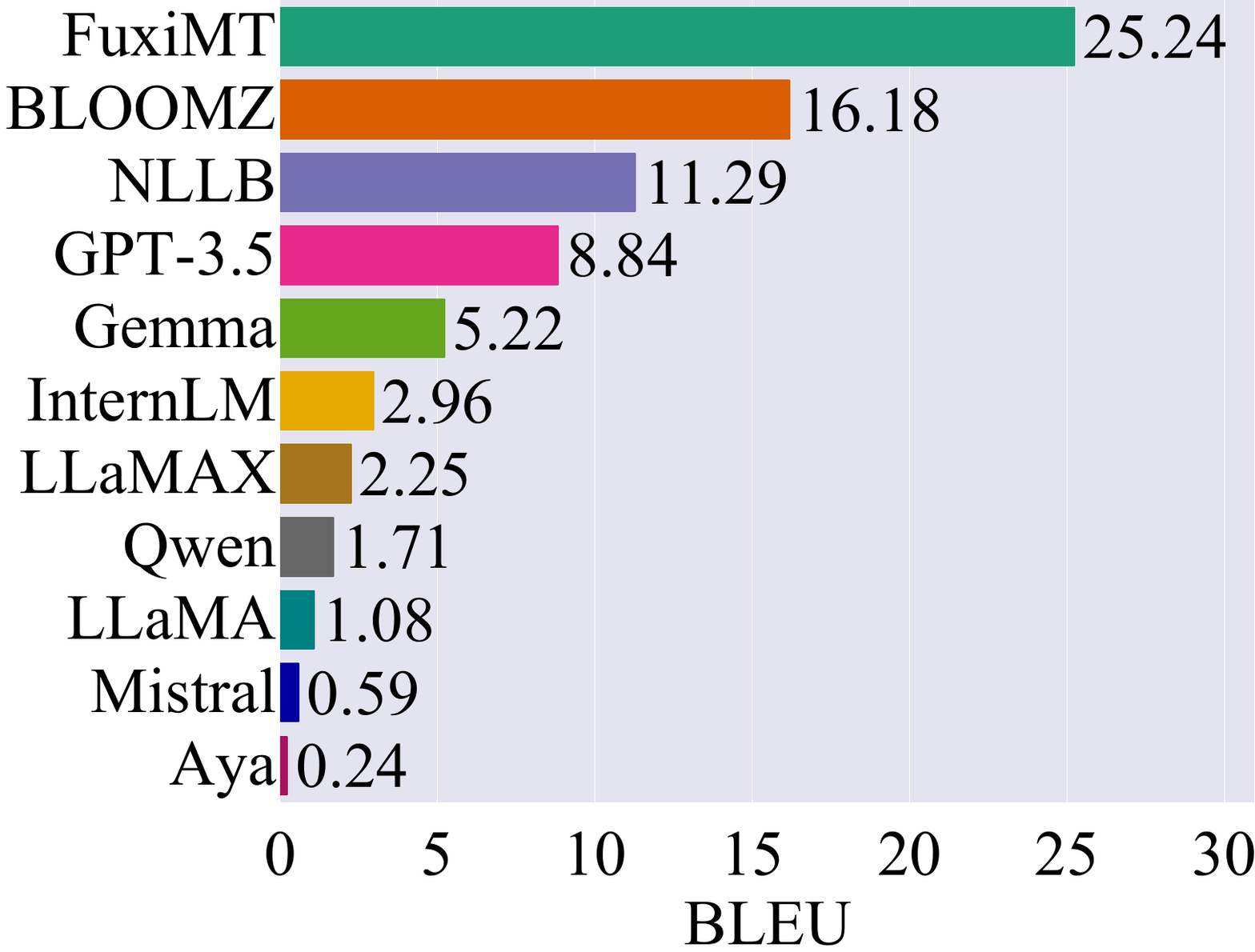}
        \caption{Tibetan}
    \end{subfigure}

    \vfill

    \begin{subfigure}[b]{0.44\textwidth}
        \includegraphics[width=\textwidth]{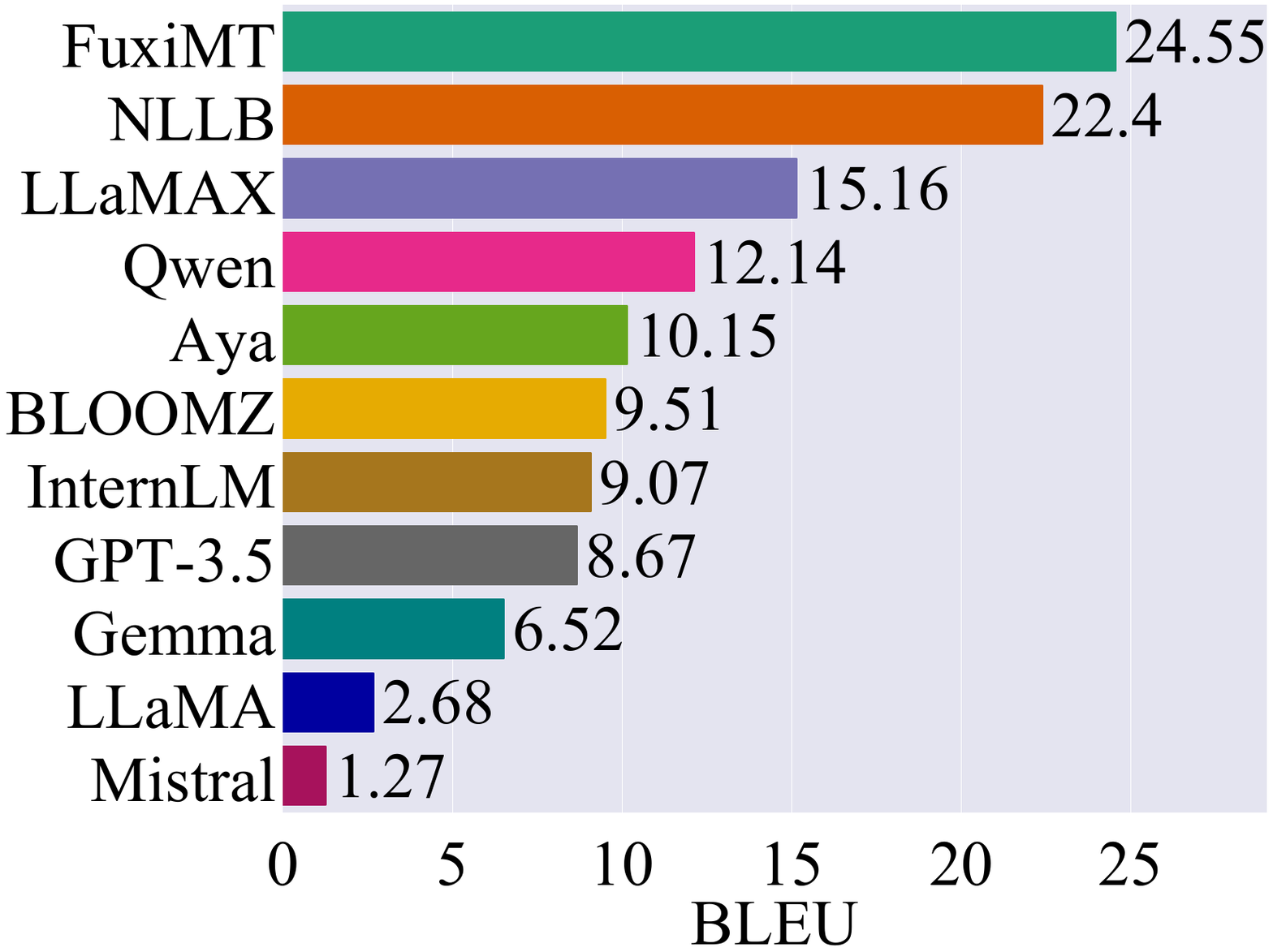}
        \caption{Turkmen}
    \end{subfigure}
    \begin{subfigure}[b]{0.44\textwidth}
        \includegraphics[width=\textwidth]{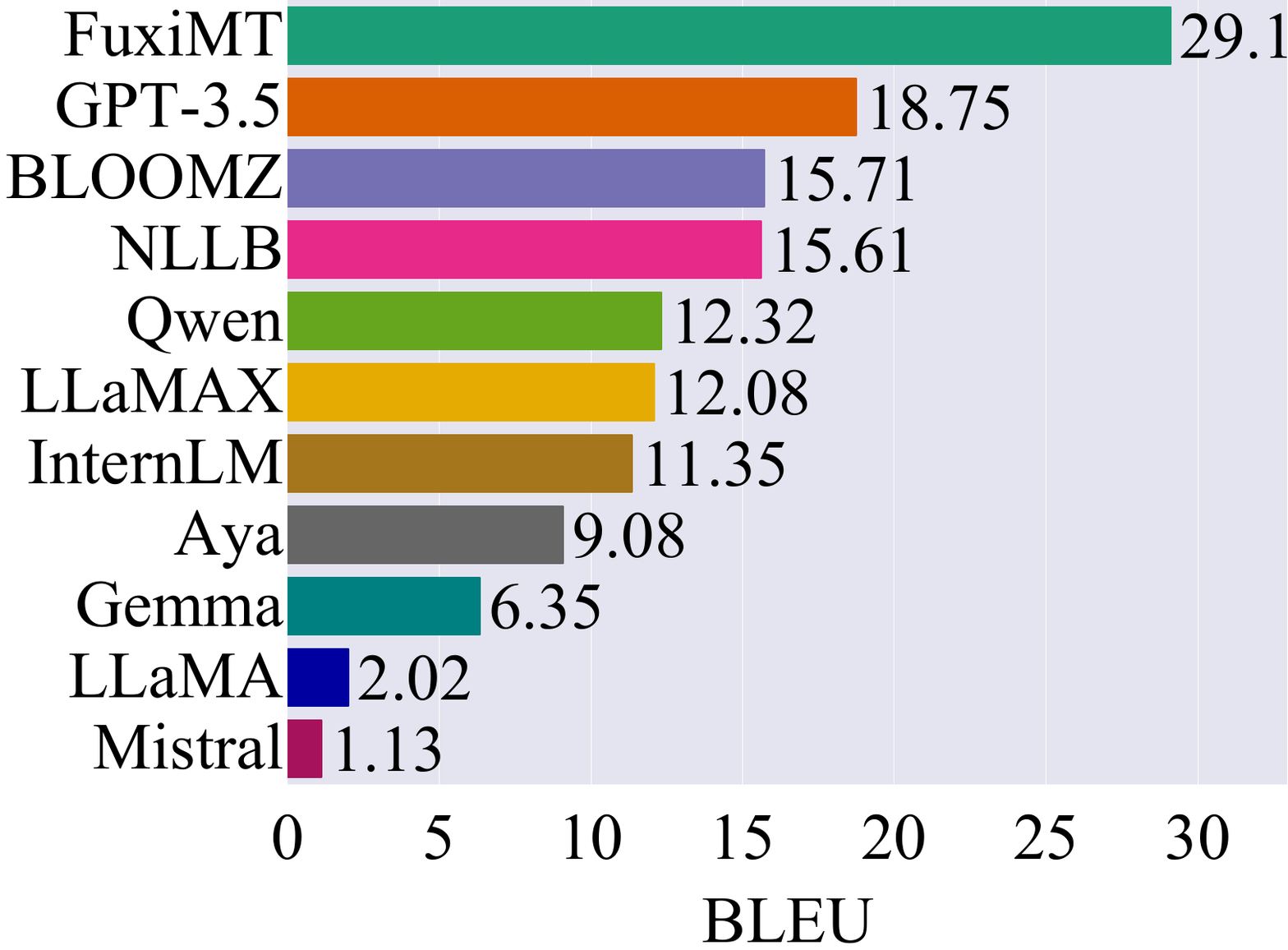}
        \caption{Pijin}
    \end{subfigure}
    \caption{Results on zero-shot language pairs}
    \label{fig-zero}
\end{figure*}

FuxiMT's generalization ability was further evaluated through zero-shot translation on four unseen language pairs: Tigrinyan (ti), Tibetan (bo), Turkmen (tk), and Pijin (pis). 
As shown in Figure \ref{fig-zero}, FuxiMT consistently outperforms all baseline models across these pairs, demonstrating its remarkable capacity to transfer knowledge to unseen languages without explicit training data. 
This success is attributed to several key factors: (1) the multilingual training with curriculum learning, which exposes the model to diverse language pairs and translation patterns, promoting robust cross-lingual representations
(2) and the Mixture-of-Experts architecture, which facilitates efficient knowledge sharing and transfer between different language pairs, even in zero-shot scenarios.

The incorporation of MoEs, coupled with the data augmentation via back-translation (applied to seen pairs), further enhances FuxiMT’s generalization capabilities, enabling it to handle unseen language pairs more effectively. 
These combined strategies demonstrate FuxiMT’s potential as a powerful tool for cross-lingual communication, especially for low-resource scenarios. The strong zero-shot performance highlights FuxiMT's ability to bridge communication gaps where parallel data is scarce or non-existent, a crucial advantage for fostering seamless communication and collaboration across the diverse linguistic landscape.

\subsection{Comparison to Other MT model based on LLM}

BigTranslate is another MT model based on LLM which demonstrates strong performance on the FLORES-200 dataset. 
In this section, we compare FuxiMT with BigTranslate on FLORES-200.
As shown in Table \ref{tab:ot}, FuxiMT consistently achieves higher BLEU scores across a range of language pairs, especially those involving Chinese. 
This advantage stems from FuxiMT's Chinese-centric pre-training, which provides a strong foundation for Chinese translation, and its multilingual training with curriculum learning, enabling effective cross-lingual knowledge transfer. 
The MoE architecture further enhances FuxiMT's ability to handle diverse linguistic nuances and low-resource scenarios, leading to overall improved translation quality compared to BigTranslate's more general LLM approach. 

\begin{table}[t]
\centering
\footnotesize
\begin{tabular}{ccc}
\toprule
Lang & FuxiMT & BigTranslate \\
\midrule
en & 53.12 & 31.1 \\
es & 46.78 & 23.5 \\
fr & 47.73 & 24.2 \\
pt & 39.47 & 23.6 \\
id & 41.58 & 18.4 \\
hi & 28.93 & 15.1 \\
ko & 30.27 & 19.6 \\
ne & 29.92 & 10.8 \\
ka & 12.22 & 0.9 \\
my & 11.32 & 5.9 \\
bo & 25.24 & 36.3 \\
\bottomrule
\end{tabular}
\caption{Comparison with BigTranslate}
\label{tab:ot}
\end{table}

\subsection{Instructions for Implement Machine Translation}
Multilingual Neural Machine Translation (MNMT) continues to evolve but faces significant challenges in balancing performance across multiple languages. Translation approaches based on Large Language Models (LLMs) each have their advantages and limitations. FuxiMT introduces an innovative solution by combining a two-stage training process with a mixture of experts (MoEs) model, effectively addressing existing issues.

In practical applications, a comprehensive set of detailed translation instructions has been developed to ensure translation quality and consistency (see Table \ref{tab:mt-instructions}). These instructions cover a variety of expressions designed to guide the model in accurately understanding the translation task, specifying the source and target languages, and addressing different text units such as sentences and phrases. For instance, instructions like “Translate from \{src\_lang\} to \{tgt\_lang\}: \{src\_text\}” help the model better grasp the translation intent. By leveraging these directives, FuxiMT produces more accurate and fluent translations across different language pairs. It excels particularly in handling low-resource language pairs and complex linguistic structures, significantly improving both accuracy and robustness. This provides an effective solution for multilingual translation challenges.
\begin{table*}[t]
\centering
\footnotesize
\begin{tabular}{cc}
\toprule
Resource Level & Languages \\
\midrule
High Resource  & en, es, fr, de, pt, it, ru, pl, cs, hu, el, ro, sk, bg, ar \\
Medium Resource & tr, sl, lt, zh, et, id, lv, vi, hr, ja, uk, fa, hi, ko, bn, sq, mk, ms, th, bs, sw, kk \\
Low Resource   & sr, ta, be, ur, ne, si, az, ka, hy, am, km, my, ky, mg, mn, so \\
Very Low Resource   & ps, ha, lo, rw, mi, ug, prs, ti, bo, tk, pis \\
\bottomrule
\end{tabular}
\caption{Resource level of different languages.}
\label{tab:language-resource}
\end{table*}


\begin{table*}[t]
\centering
\resizebox{0.95\textwidth}{!}{
\begin{tabular}{lll|lll|lll} 
\toprule
ISO-639 & Language & Language Family & ISO-639 & Language & Language Family &  ISO-639 &  Language & Language Family \\
\midrule
am & Amharic & Afro-Asiatic &  fr & French & Indo-European & ro & Romanian & Indo-European \\
ar & Arabic & Afro-Asiatic & ka & Georgian & Kartvelian  & ru & Russian & Indo-European \\
hy & Armenian & Indo-European & de & German & Indo-European & rw & Rwandan & Niger-Congo \\
az & Azerbaijani & Turkic & el & Greek & Indo-European & si & Sinhalese & Indo-European \\
be & Belarusian & Indo-European & ha & Hausa & Afro-Asiatic & sk & Slovak & Indo-European \\
bn & Bengali & Indo-European  & hi & Hindi & Indo-European & sl & Slovenian & Indo-European \\
bo & Tibetan & Sino-Tibetan & hr & Croatian & Indo-European &  so & Somali & Afro-Asiatic \\
bs & Bosnian & Indo-European & hu & Hungarian & Uralic & es & Spanish & Indo-European \\
bg & Bulgarian & Indo-European & id & Indonesian & Austronesian &  sw & Swahili & Niger-Congo \\
my & Burmese & Sino-Tibetan & it & Italian & Indo-European & ta & Tamil & Dravidian \\
km & Cambodian & Austroasiatic & ja & Japanese & Japonic & th & Thai & Tai-Kadai \\
cs & Czech & Indo-European &  kk & Kazakh & Turkic & ti & Tigranian & Afro-Asiatic \\
prs & Dari & Indo-European & ko & Korean & Koreanic & tr & Turkish & Turkic \\
en & English & Indo-European &  ky & Kyrgyz & Turkic & tk & Turkmen & Turkic \\
et & Estonian & Uralic &  lo & Laotian & Tai-Kadai & uk & Ukrainian & Indo-European \\
fa & Persian & Indo-European &  lt & Lithuanian & Indo-European & ur & Urdu & Indo-European \\
pis & Pijin & English Creole & mg & Madagascar & Austronesian &  ug & Uyghur & Turkic \\
pl & Polish & Indo-European & mi & Maori & Austronesian &  vi & Vietnamese & Austroasiatic \\
pt & Portuguese & Indo-European &  mk & Macedonian & Indo-European & zh & Chinese & Sino-Tibetan \\ 
mn & Mongolian & Mongolic & ms & Malay & Austronesian & sq & Albanian & Indo-European \\
ne & Nepali & Indo-European &  ps & Pashto & Indo-European & lv & Latvian & Indo-European\\
sr & Serbian & Indo-European & ro & Moldovan & Indo-European &&&\\
\bottomrule
\end{tabular}
}
\caption{The list of 65 natural languages supported by FuxiMT.}
\label{tab:language-codes}
\end{table*}

\begin{table*}[t]
\centering
\footnotesize
\begin{tabular}{cccccccccccc}
\toprule
Lang & FuxiMT & GPT-3.5 & BLOOMz & NLLB & LLaMAX3 & LLaMA3 & Mistral & Qwen & Aya & Gemma & InternLM \\
\midrule
en & 53.12 & 47.76 & 21.29 & 25.93 & 30.33 & 7.69 & 8.80 & 40.21 & 34.78 & 39.99 & 42.51 \\
es & 46.78 & 35.84 & 18.81 & 17.74 & 26.17 & 2.85 & 1.95 & 30.48 & 28.37 & 29.25 & 29.68 \\
fr & 47.73 & 28.44 & 20.64 & 2.97 & 29.49 & 1.47 & 2.06 & 34.75 & 31.37 & 12.67 & 34.28 \\
de & 43.05 & 28.48 & 12.31 & 19.83 & 29.36 & 1.40 & 1.80 & 34.10 & 31.01 & 18.86 & 33.58 \\
pt & 39.47 & 34.11 & 19.29 & 19.01 & 28.77 & 2.30 & 1.43 & 34.44 & 31.78 & 5.37 & 34.49 \\
it & 38.59 & 32.64 & 11.09 & 18.08 & 26.93 & 2.30 & 1.95 & 31.83 & 29.59 & 4.10 & 30.39 \\
ru & 36.63 & 25.89 & 15.75 & 20.55 & 28.03 & 1.48 & 3.76 & 32.45 & 29.29 & 30.49 & 30.96 \\
pl & 36.30 & 27.20 & 5.82 & 15.68 & 25.91 & 4.68 & 1.31 & 29.23 & 27.84 & 16.81 & 26.53 \\
cs & 35.69 & 22.26 & 6.13 & 18.26 & 28.14 & 2.91 & 1.91 & 31.91 & 30.87 & 18.44 & 30.13 \\
hu & 29.95 & 19.29 & 9.07 & 18.50 & 27.15 & 2.31 & 1.83 & 26.18 & 19.15 & 12.32 & 23.58 \\
el & 28.92 & 17.47 & 11.91 & 19.58 & 26.64 & 1.52 & 14.54 & 25.80 & 28.23 & 4.09 & 20.22 \\
ro & 30.05 & 27.06 & 9.63 & 22.09 & 29.76 & 2.79 & 4.48 & 32.23 & 31.18 & 32.41 & 32.17 \\
sk & 24.68 & 19.18 & 8.62 & 17.12 & 27.99 & 2.16 & 2.07 & 29.30 & 27.15 & 29.81 & 25.99 \\
bg & 26.41 & 20.82 & 21.29 & 18.88 & 28.63 & 1.37 & 3.59 & 31.10 & 24.29 & 31.81 & 28.89 \\
ar & 38.01 & 22.25 & 13.46 & 20.92 & 27.34 & 1.41 & 2.21 & 31.20 & 29.65 & 20.68 & 20.78 \\
tr & 34.72 & 25.18 & 7.46 & 18.97 & 26.12 & 1.73 & 4.63 & 28.90 & 26.88 & 4.99 & 27.41 \\
sl & 24.93 & 23.19 & 10.23 & 17.60 & 26.73 & 2.30 & 1.37 & 26.89 & 19.80 & 3.43 & 23.67 \\
lt & 15.45 & 14.18 & 9.42 & 17.12 & 26.33 & 3.82 & 1.29 & 24.08 & 20.58 & 15.11 & 16.54 \\
et & 23.14 & 17.35 & 9.50 & 17.71 & 27.28 & 1.27 & 1.13 & 24.12 & 13.53 & 13.11 & 17.10 \\
id & 41.58 & 37.45 & 18.66 & 19.65 & 28.46 & 1.40 & 1.43 & 33.83 & 30.24 & 31.46 & 29.57 \\
lv & 12.74 & 10.85 & 10.95 & 16.81 & 28.09 & 2.18 & 9.68 & 25.79 & 15.18 & 30.06 & 17.93 \\
vi & 23.85 & 21.84 & 17.05 & 21.13 & 27.05 & 6.22 & 4.47 & 32.19 & 29.35 & 27.91 & 24.37 \\
hr & 28.47 & 27.87 & 7.98 & 17.04 & 27.59 & 2.25 & 1.47 & 29.44 & 24.61 & 28.27 & 27.21 \\
ja & 33.05 & 23.79 & 13.59 & 17.15 & 22.69 & 1.32 & 17.99 & 28.97 & 26.89 & 16.86 & 27.75 \\
uk & 23.14 & 21.13 & 18.22 & 21.06 & 25.93 & 1.31 & 4.69 & 30.33 & 29.53 & 14.41 & 28.46 \\
fa & 23.19 & 16.60 & 12.14 & 21.90 & 24.66 & 1.43 & 7.96 & 28.12 & 28.24 & 6.59 & 19.97 \\
hi & 28.93 & 19.49 & 14.72 & 20.81 & 25.86 & 1.28 & 12.65 & 26.97 & 25.64 & 4.05 & 19.13 \\
ko & 30.27 & 25.47 & 5.39 & 22.16 & 24.27 & 1.46 & 13.91 & 29.59 & 27.05 & 7.73 & 24.99 \\
bn & 26.68 & 14.07 & 14.47 & 19.89 & 24.56 & 1.35 & 10.35 & 24.35 & 8.43 & 26.56 & 12.40 \\
sq & 21.17 & 17.21 & 9.93 & 19.44 & 27.03 & 2.52 & 4.43 & 18.44 & 13.10 & 28.12 & 14.20 \\
mk & 12.84 & 10.48 & 16.39 & 19.59 & 28.98 & 1.33 & 4.95 & 29.34 & 18.97 & 6.09 & 24.73 \\
ms & 36.03 & 33.60 & 16.58 & 22.39 & 27.61 & 1.29 & 4.39 & 31.05 & 26.14 & 11.05 & 26.32 \\
th & 22.74 & 17.75 & 11.65 & 20.13 & 25.34 & 1.45 & 12.49 & 29.09 & 13.86 & 23.76 & 16.48 \\
bs & 27.82 & 18.71 & 12.95 & 18.76 & 29.14 & 1.69 & 1.84 & 30.35 & 24.19 & 30.73 & 28.36 \\
sw & 25.01 & 16.48 & 10.11 & 18.25 & 24.60 & 1.37 & 1.27 & 12.07 & 6.91 & 3.07 & 8.88 \\
kk & 14.88 & 7.54 & 13.04 & 21.48 & 26.18 & 1.49 & 5.10 & 17.21 & 7.50 & 3.34 & 9.13 \\
sr & 28.04 & 16.94 & 10.97 & 23.34 & 29.45 & 1.34 & 2.07 & 29.98 & 19.39 & 2.66 & 24.21 \\
ta & 19.53 & 13.07 & 10.81 & 20.92 & 22.70 & 1.31 & 3.94 & 13.37 & 10.74 & 4.50 & 4.47 \\
be & 27.73 & 7.95 & 19.05 & 18.64 & 22.92 & 1.49 & 1.99 & 20.97 & 13.84 & 15.43 & 16.91 \\
ur & 25.27 & 16.49 & 9.24 & 22.20 & 23.61 & 1.41 & 9.25 & 21.40 & 10.28 & 2.99 & 13.23 \\
ne & 29.92 & 14.84 & 15.72 & 21.80 & 25.39 & 1.29 & 1.42 & 20.22 & 12.22 & 8.83 & 11.22 \\
si & 21.31 & 11.16 & 11.10 & 19.99 & 20.97 & 1.40 & 0.71 & 8.30 & 1.42 & 8.26 & 2.05 \\
az & 27.88 & 13.00 & 8.51 & 13.95 & 11.68 & 1.33 & 3.08 & 9.13 & 8.63 & 2.15 & 5.30 \\
ka & 12.22 & 9.66 & 19.27 & 20.74 & 24.45 & 5.17 & 1.79 & 15.94 & 5.84 & 17.84 & 3.81 \\
hy & 25.16 & 6.15 & 20.86 & 22.11 & 28.25 & 5.30 & 4.88 & 16.41 & 4.51 & 3.51 & 3.06 \\
am & 18.02 & 5.70 & 12.63 & 22.93 & 19.08 & 5.03 & 1.03 & 4.26 & 0.78 & 16.97 & 1.49 \\
km & 11.91 & 8.52 & 18.74 & 22.29 & 22.66 & 12.63 & 1.51 & 13.30 & 0.94 & 17.64 & 4.48 \\
my & 11.32 & 5.84 & 14.09 & 16.09 & 20.36 & 5.35 & 1.81 & 7.46 & 0.74 & 17.93 & 2.01 \\
ky & 15.55 & 8.20 & 17.18 & 15.98 & 21.38 & 1.48 & 1.69 & 11.57 & 8.62 & 4.18 & 7.28 \\
mg & 17.36 & 8.54 & 14.91 & 14.09 & 17.37 & 1.25 & 0.70 & 6.12 & 3.99 & 7.94 & 5.80 \\
mn & 24.66 & 12.37 & 15.54 & 17.23 & 23.01 & 1.31 & 1.84 & 10.93 & 3.49 & 1.79 & 4.98 \\
so & 14.42 & 7.97 & 5.45 & 16.98 & 17.42 & 1.28 & 0.74 & 5.92 & 4.81 & 9.46 & 5.20 \\
ps & 18.24 & 5.16 & 10.23 & 22.90 & 21.94 & 1.44 & 2.71 & 11.27 & 4.96 & 11.38 & 5.54 \\
ha & 17.79 & 3.59 & 6.45 & 19.59 & 19.92 & 1.32 & 0.75 & 6.00 & 3.67 & 18.19 & 5.68 \\
lo & 12.33 & 6.72 & 11.30 & 23.78 & 19.84 & 4.19 & 0.72 & 9.35 & 2.55 & 1.78 & 5.08 \\
rw & 12.39 & 6.42 & 7.36 & 19.51 & 14.88 & 1.70 & 0.92 & 5.52 & 3.90 & 14.60 & 4.91 \\
mi & 24.90 & 11.15 & 12.59 & 19.88 & 16.54 & 1.30 & 1.07 & 9.62 & 4.89 & 2.52 & 7.48 \\
ug & 21.20 & 9.78 & 10.47 & 21.71 & 16.26 & 12.30 & 2.88 & 10.87 & 3.14 & 1.20 & 12.77 \\
prs & 19.89 & 11.96 & 19.27 & 26.22 & 22.41 & 1.43 & 1.42 & 26.31 & 26.35 & 1.45 & 19.01 \\
ti & 21.67 & 9.60 & 18.87 & 18.35 & 5.86 & 3.22 & 0.58 & 2.86 & 0.45 & 3.01 & 1.48 \\
bo & 25.24 & 8.84 & 16.18 & 11.29 & 2.25 & 1.08 & 0.59 & 1.71 & 0.24 & 5.22 & 2.96 \\
tk & 24.55 & 8.67 & 9.51 & 22.40 & 15.16 & 2.68 & 1.27 & 12.14 & 10.15 & 6.52 & 9.07 \\
pis & 29.10 & 18.75 & 15.71 & 15.61 & 12.08 & 2.02 & 1.13 & 12.32 & 9.08 & 6.35 & 11.35 \\
ro & 30.05 & 27.06 & 9.63 & 22.09 & 29.76 & 2.79 & 4.48 & 32.23 & 31.18 & 32.41 & 32.17 \\
\bottomrule
\end{tabular}
\caption{The detail BLEU scores in different language pairs xx-zh}
\label{bleu}
\end{table*}

\begin{table*}[t]
\centering
\footnotesize
\begin{tabular}{ccccccccccc}
\toprule
Lang  &  FuxiMT  &  GPT-3.5  &  BLOOMz  &  LLaMAX3  &  LLaMA3  &  Mistral  &  Qwen  &  Aya  &  Gemma  &  InternLM \\
\midrule
en  & 49.6 & 39.88 & 17.38 & 28.96 & 8.05 & 9.55 & 37.59 & 32.49 & 37.03 & 39.14 \\
es  & 45.68 & 29.57 & 15.82 & 25.71 & 3.62 & 2.83 & 29.44 & 27.24 & 28.53 & 28.58 \\
fr  & 46.49 & 24.21 & 16.87 & 28.26 & 2.38 & 3.05 & 32.74 & 29.42 & 13.35 & 32 \\
de  & 41.99 & 25.68 & 11.37 & 28.44 & 2.21 & 2.63 & 32.14 & 29.16 & 18.91 & 31.54 \\
pt  & 37.52 & 27.95 & 15.8 & 27.87 & 3.12 & 2.44 & 32.56 & 29.97 & 6.59 & 32.17 \\
it  & 36.08 & 27.53 & 10.6 & 26.36 & 3.14 & 2.98 & 30.48 & 28.2 & 5.09 & 29.04 \\
ru  & 34.34 & 19.78 & 13.42 & 26.73 & 2.21 & 4.85 & 30.76 & 27.71 & 28.95 & 29.3 \\
pl  & 34.34 & 22.48 & 6.48 & 25.7 & 5.49 & 2.26 & 28.42 & 27.01 & 17.46 & 25.7 \\
cs  & 34.67 & 17.86 & 6.58 & 27.41 & 3.83 & 2.85 & 30.43 & 29.38 & 18.8 & 28.64 \\
hu  & 31.42 & 16.58 & 8.68 & 26.56 & 3.07 & 2.86 & 26.01 & 19.71 & 13.09 & 23.36 \\
el  & 29.12 & 13.51 & 10.92 & 25.83 & 2.41 & 15.87 & 25.22 & 26.93 & 4.9 & 20.43 \\
ro  & 29.46 & 23.14 & 9.06 & 28.7 & 3.67 & 5.59 & 30.73 & 29.33 & 31.26 & 30.44 \\
sk  & 23.54 & 15.39 & 8.44 & 27.47 & 2.97 & 3.06 & 28.37 & 26.36 & 29.25 & 25.38 \\
bg  & 24.69 & 16.34 & 17.31 & 27.16 & 2.15 & 4.65 & 29.52 & 23.76 & 30.21 & 27.47 \\
ar  & 35.38 & 15.53 & 12.13 & 26.07 & 2.28 & 3.3 & 29.55 & 28.12 & 20.11 & 20.71 \\
tr  & 32.78 & 20.53 & 7.5 & 26.06 & 2.49 & 6.14 & 28.26 & 26.37 & 6.27 & 26.56 \\
sl  & 23.67 & 19.32 & 9.66 & 26.27 & 3.09 & 2.35 & 26.28 & 20.3 & 4.75 & 23.19 \\
lt  & 14.02 & 13.47 & 8.87 & 25.58 & 4.71 & 2.25 & 23.81 & 20.73 & 15.57 & 17.24 \\
et  & 22.35 & 15.04 & 9.32 & 26.75 & 2.05 & 1.98 & 24 & 14.95 & 13.93 & 17.87 \\
id  & 39.89 & 30.59 & 15.75 & 27.62 & 2.25 & 2.28 & 32 & 28.77 & 30.41 & 28.15 \\
lv  & 13.73 & 9.91 & 10.34 & 27.1 & 3.1 & 11.54 & 25.24 & 16.29 & 28.97 & 18.55 \\
vi  & 19.85 & 19.08 & 14.2 & 26.68 & 6.96 & 5.79 & 30.87 & 28.14 & 27.55 & 24.2 \\
hr  & 27.05 & 22.9 & 7.82 & 26.97 & 3.12 & 2.33 & 28.38 & 24.16 & 27.67 & 26.11 \\
ja  & 38.28 & 22.92 & 11.99 & 22.41 & 2.08 & 18.88 & 27.91 & 26.01 & 16.95 & 26.65 \\
uk  & 22.29 & 16.35 & 15.2 & 25.13 & 2.04 & 5.61 & 28.92 & 27.94 & 14.79 & 27.2 \\
fa  & 22.44 & 11.74 & 11.19 & 24.48 & 2.29 & 9.61 & 27.41 & 27.16 & 7.51 & 20.31 \\
hi  & 26.2 & 16.05 & 12.86 & 25.14 & 2.04 & 14.3 & 26.09 & 24.73 & 4.7 & 19.46 \\
ko  & 32.39 & 20.05 & 6.3 & 24.06 & 2.16 & 15.07 & 28.54 & 26.44 & 8.71 & 24.61 \\
bn  & 25.09 & 10.93 & 12.62 & 23.99 & 2.19 & 12.17 & 23.98 & 10.41 & 25.66 & 13.76 \\
sq  & 20.92 & 16.91 & 9.16 & 26.55 & 3.42 & 6.04 & 19.12 & 14.43 & 27.78 & 15.36 \\
mk  & 12.29 & 9.38 & 14.18 & 27.59 & 2.07 & 6.21 & 28.02 & 19.3 & 7.13 & 23.83 \\
ms  & 35.71 & 28.85 & 14.51 & 27.09 & 2 & 5.46 & 29.81 & 25.41 & 12.1 & 25.5 \\
th  & 24.57 & 13.75 & 11.26 & 25.12 & 2.07 & 14.46 & 28.19 & 15.52 & 23.5 & 17.34 \\
bs  & 25.46 & 17.07 & 11.59 & 28.13 & 2.56 & 2.8 & 29.06 & 23.65 & 29.78 & 27.12 \\
sw  & 23.75 & 13.67 & 9.89 & 24.43 & 2.23 & 2.16 & 13.69 & 8.53 & 4.14 & 10.3 \\
kk  & 15.82 & 6.42 & 11.78 & 25.32 & 2.24 & 6.97 & 18.04 & 9.03 & 4.31 & 10.64 \\
sr  & 25.49 & 14 & 10.13 & 28.02 & 2.12 & 2.92 & 28.4 & 19.6 & 3.64 & 23.54 \\
ta  & 20.21 & 9.3 & 10.29 & 22.36 & 2.11 & 5.94 & 14.65 & 12.3 & 5.77 & 6.45 \\
be  & 28.54 & 7.86 & 15.95 & 22.95 & 2.42 & 3.21 & 21.65 & 15.33 & 15.99 & 18.06 \\
ur  & 23.58 & 12.63 & 9.3 & 23.27 & 2.21 & 11.17 & 21.46 & 11.83 & 4.02 & 14.41 \\
ne  & 30.08 & 12.85 & 13.54 & 24.75 & 2.05 & 2.48 & 20.51 & 13.44 & 9.71 & 12.85 \\
si  & 23.06 & 8.1 & 10.12 & 21.01 & 2.24 & 1.31 & 10.05 & 2.83 & 9.24 & 3.78 \\
az  & 28.66 & 10 & 8.1 & 12.92 & 2.1 & 4.5 & 10.82 & 10.08 & 2.88 & 7.05 \\
ka  & 13.68 & 7.06 & 15.99 & 23.95 & 5.98 & 2.92 & 16.88 & 7.96 & 17.81 & 5.92 \\
hy  & 26.22 & 5.12 & 16.93 & 26.78 & 5.97 & 6.57 & 17.07 & 6.52 & 4.34 & 4.91 \\
am  & 19.08 & 5.45 & 11.63 & 19.33 & 6.3 & 2.08 & 6.09 & 1.48 & 17.59 & 3.19 \\
km  & 13.3 & 6.8 & 15.87 & 23.52 & 14.3 & 2.91 & 15.92 & 1.88 & 19.49 & 6.94 \\
my  & 12.89 & 5.47 & 13.04 & 20.63 & 6.47 & 2.99 & 9.64 & 1.33 & 18.76 & 3.46 \\
ky  & 16.29 & 5.76 & 14.45 & 21.47 & 2.26 & 2.71 & 13.35 & 10.38 & 5.15 & 8.99 \\
mg  & 18.39 & 9.29 & 13.12 & 18.39 & 2.09 & 1.21 & 8.1 & 5.44 & 9.62 & 7.59 \\
mn  & 25.35 & 10.63 & 13.28 & 22.75 & 2.09 & 2.81 & 12.45 & 4.71 & 2.67 & 6.68 \\
so  & 15.47 & 7.55 & 6.49 & 18.35 & 2.01 & 1.29 & 7.79 & 6.47 & 10.95 & 6.72 \\
ps  & 18.52 & 3.73 & 10 & 22.41 & 2.3 & 4.21 & 13.03 & 6.48 & 12.71 & 7.38 \\
ha  & 18.33 & 3.12 & 7.4 & 20.58 & 2.16 & 1.24 & 7.82 & 5.04 & 19.46 & 7.18 \\
lo  & 14.19 & 5.77 & 10.62 & 20.8 & 5.89 & 1.34 & 11.97 & 4.21 & 2.92 & 7.19 \\
rw  & 14.12 & 5.8 & 7.46 & 16.25 & 2.49 & 1.59 & 7.44 & 5.26 & 16.47 & 6.5 \\
mi  & 26.47 & 10.1 & 11.41 & 18.01 & 2.11 & 1.92 & 11.7 & 6.46 & 3.54 & 9.31 \\
ug  & 22.27 & 8.47 & 9.9 & 17.04 & 13.01 & 4.49 & 12.51 & 4.55 & 2.01 & 14.18 \\
prs  & 20.09 & 8.86 & 16.02 & 22.36 & 2.28 & 2.41 & 25.65 & 25.4 & 2.25 & 19.36 \\
ti  & 23.11 & 7.02 & 15.93 & 7.32 & 4.79 & 1.24 & 4.53 & 1.04 & 4.03 & 2.92 \\
bo  & 26.56 & 5.8 & 13.86 & 3.79 & 1.98 & 1.17 & 3.33 & 0.74 & 6.87 & 5.19 \\
tk  & 26.05 & 5.66 & 9 & 16.1 & 3.63 & 2.19 & 13.65 & 11.71 & 7.5 & 10.69 \\
pis  & 25.48 & 17.22 & 13.38 & 13.57 & 3.08 & 2.06 & 14.28 & 10.72 & 7.83 & 13.17 \\
ro  & 29.46 & 23.14 & 9.06 & 28.7 & 3.67 & 5.59 & 30.73 & 29.33 & 31.26 & 30.44 \\
\bottomrule
\end{tabular}
\caption{The detail chrf scores in different language pairs xx-zh}
\label{chrf}
\end{table*}

\begin{table*}[t]
\centering
\small
\begin{tabular}{l}
\toprule
\textbf{Instructions} \\
\midrule
How do you say \{src\_text\} in \{tgt\_lang\}?\\
\{src\_text\} How do you say this sentence in \{tgt\_lang\}?\\
\{src\_text\} Say this using \{tgt\_lang\}\\
Translate from \{src\_lang\} to \{tgt\_lang\}: \{src\_text\}\\
Translate \{src\_text\} from \{src\_lang\} to \{tgt\_lang\}.\\
Translate \{src\_text\} to \{tgt\_lang\}.\\
Translate the following. \{src\_lang\}: \{src\_text\} \{tgt\_lang\}:\\
Translate the sentence from \{src\_lang\} to \{tgt\_lang\}. \{src\_lang\}: \{src\_text\} Corresponding \{tgt\_lang\} translation: \\
Translate the sentence from \{src\_lang\} to \{tgt\_lang\}. \{src\_lang\}: \{src\_text\} \{tgt\_lang\}: \\
Translate from \{src\_lang\} to \{tgt\_lang\}. \{src\_lang\}: \{src\_text\} \{tgt\_lang\}: \\
Render the \{src\_lang\} sentence \{src\_text\} into \{tgt\_lang\}.\\
Produce the \{tgt\_lang\} equivalent for the \{src\_lang\} phrase \{src\_text\}.\\
Present \{src\_text\} in \{tgt\_lang\}, originally written in \{src\_lang\}.\\
Translate the \{src\_lang\} text \{src\_text\} to \{tgt\_lang\}.\\
Provide a \{tgt\_lang\} version of this \{src\_lang\} statement: \{src\_text\}\\
Reword \{src\_text\} from \{src\_lang\} into \{tgt\_lang\}.\\
Convert this \{src\_lang\} expression \{src\_text\} to \{tgt\_lang\}.\\
Express the \{src\_lang\} phrase \{src\_text\} in \{tgt\_lang\}.\\
How would you phrase \{src\_text\} in \{tgt\_lang\}?\\
Transcribe \{src\_text\} from \{src\_lang\} to \{tgt\_lang\}.\\
Can you adapt \{src\_text\}, which is in \{src\_lang\}, to \{tgt\_lang\}?\\
Rephrase \{src\_text\} in \{tgt\_lang\}, its original language is \{src\_lang\}.\\
Show me \{src\_text\} written in \{tgt\_lang\} instead of \{src\_lang\}.\\
Please rewrite \{src\_text\} in \{tgt\_lang\}, it was originally in \{src\_lang\}.\\
Represent \{src\_text\} in \{tgt\_lang\}, its source language is \{src\_lang\}.\\
Translate \{src\_text\} which is in \{src\_lang\}, to \{tgt\_lang\}.\\
How can \{src\_text\} be expressed in \{tgt\_lang\}?\\
Give me the \{tgt\_lang\} translation for \{src\_text\}.\\
I need \{src\_text\} written in \{tgt\_lang\}, please help.\\
Please provide the \{tgt\_lang\} translation for the following \{src\_lang\} sentence: \{src\_text\}\\
Convert the \{src\_lang\} sentence \{src\_text\} into \{tgt\_lang\}.\\
What is the \{tgt\_lang\} version of the \{src\_lang\} sentence \{src\_text\}?\\
I need the \{tgt\_lang\} rendition of this \{src\_lang\} phrase: \{src\_text\}\\
Transform \{src\_text\} from \{src\_lang\} to \{tgt\_lang\}.\\
Translate \{src\_text\} from \{src\_lang\} to \{tgt\_lang\} for me.\\
Can you change \{src\_text\} which is in \{src\_lang\} to \{tgt\_lang\}?\\
Give me the \{tgt\_lang\} equivalent of this \{src\_lang\} sentence: \{src\_text\}\\
Rewrite the \{src\_lang\} phrase \{src\_text\} in \{tgt\_lang\}.\\
I would like the \{tgt\_lang\} translation of the following \{src\_lang\} sentence, please: \{src\_text\}\\
\bottomrule
\end{tabular}
\caption{Examples of translation instruction}
\label{tab:mt-instructions}
\end{table*}

\subsection{Related Work}
\label{Related Work}

This section delves into the existing landscape of multilingual machine translation and the recent advancements in leveraging large language models for this task, highlighting the limitations of current approaches and positioning FuxiMT as a novel solution.

\subsubsection{Multilingual Neural Machine Translation}

Multilingual Neural Machine Translation (MNMT) has made significant strides in recent years, aiming to build a single system capable of translating between multiple language pairs \cite{DBLP:conf/acl/ChengBF0WM22}. 
This approach offers significant advantages over training individual bilingual systems, particularly in low-resource scenarios where parallel data is limited.
MNMT leverages the power of cross-lingual transfer learning, allowing knowledge acquired from high-resource language pairs to enhance the translation quality for low-resource pairs \cite{DBLP:journals/tacl/JohnsonSLKWCTVW17}. 
Early MNMT research focused on extending traditional statistical machine translation methods to handle multiple languages \cite{DBLP:journals/coling/DurraniSFKS15}. 
However, the emergence of neural machine translation (NMT) revolutionized the field, leading to significant improvements in translation quality. 
Neural MNMT models typically employ encoder-decoder architectures, where the encoder maps the source sentence into a context vector and the decoder generates the target sentence based on this context. 
Numerous techniques have been proposed to further enhance the performance of neural MNMT, including parameter sharing \cite{DBLP:conf/naacl/FiratCB16}, language-specific components\cite{DBLP:conf/acl/LinWWL20}, and data augmentation \cite{DBLP:journals/corr/abs-2303-15265}. 
These techniques aim to encourage cross-lingual transfer learning, capture language-specific nuances, and address the data scarcity issue, respectively. 
Despite these advancements, achieving balanced performance across multiple language pairs, especially for those with limited resources, remains a significant challenge in MNMT.

\subsubsection{Large Language Models for Machine Translation}

The emergence of LLMs has opened up new possibilities for multilingual translation. 
These models, trained on vast amounts of text data, exhibit remarkable capabilities in understanding and generating natural language, even in zero-shot or few-shot settings. Several research directions have explored adapting LLMs for MT \cite{DBLP:journals/corr/abs-2306-10968,yang2023bigtranslate}. 
Direct fine-tuning, a straightforward approach, involves training pre-trained LLMs on parallel data for the specific task of translation. 
However, this method can suffer from catastrophic forgetting, where the model loses previously acquired linguistic knowledge during fine-tuning \cite{DBLP:conf/emnlp/ZhuFZWL21,DBLP:conf/aaai/0002C21}. 
Parameter-efficient fine-tuning methods like Adapters \cite{DBLP:conf/icml/HoulsbyGJMLGAG19} and \cite{DBLP:conf/iclr/HuSWALWWC22} aim to reduce computational cost and mitigate catastrophic forgetting by fine-tuning only a small subset of the LLM's parameters. 
While offering efficiency and stability, these techniques may not fully exploit the vast linguistic knowledge already encoded within the LLM, potentially limiting its cross-lingual transfer learning capabilities \cite{DBLP:conf/iclr/Xu0SA24}. 
Prompt engineering and in-context learning leverage the model's ability to learn from a few examples provided during inference, eliminating the need for explicit fine-tuning \cite{DBLP:conf/coling/ZhuCX24,DBLP:journals/ipm/ZhuPX24}. 
However, achieving consistent and high-quality translation across diverse language pairs using these methods remains challenging, as the quality heavily relies on the design of prompts and the selection of in-context examples, making it difficult to generalize across different language pairs and domains \cite{DBLP:journals/corr/abs-2402-15061}. 
Finally, Mixture-of-Experts (MoEs) enhance the capacity and efficiency of LLMs by allowing specialized experts to handle different aspects of the input space. 
Models like GLaM \cite{DBLP:conf/icml/DuHDTLXKZYFZFBZ22} and PanGu-$\Sigma$ \cite{DBLP:journals/corr/abs-2303-10845} demonstrate the effectiveness of MoEs in scaling LLMs and achieving state-of-the-art results on various NLP tasks, including MT. 
However, these models are typically trained from scratch with MoEs and do not explicitly address the challenge of adapting existing pre-trained multilingual LLMs for MT. 
This approach may miss out on leveraging existing linguistic knowledge and requires significant computational resources for training from scratch.

\subsubsection{Bridging the Gap: FuxiMT}

Our work, FuxiMT, builds upon the advancements in both MNMT and LLM-based translation, drawing inspiration from the FuxiTranyu paper \cite{DBLP:journals/corr/abs-2408-06273} and addressing the limitations of existing approaches. 
Unlike previous work that focuses on either direct fine-tuning, parameter-efficient methods, or prompting, FuxiMT leverages MoEs for efficient knowledge transfer while preserving the pre-trained BLOOM model's core linguistic capabilities. 
This approach mitigates the risk of catastrophic forgetting, allowing FuxiMT to retain the benefits of pre-training while acquiring translation-specific expertise. 
Similar to FuxiTranyu, FuxiMT adopts a two-stage training strategy, first establishing a Chinese-centric model through pre-training on a massive Chinese corpus and then enhancing its multilingual capabilities using a large-scale parallel corpus.
However, FuxiMT distinguishes itself by uniquely incorporating MoEs to enhance the model's ability to handle diverse language pairs. 
This novel combination of techniques positions FuxiMT as a promising solution for achieving high-quality, Chinese-centric multilingual translation, particularly for under-resourced languages.

\end{document}